\pdfoutput=1

\documentclass[11pt]{article}

\usepackage{acl}

\usepackage{times}
\usepackage{latexsym}
\usepackage{float}
\usepackage{subcaption}
\usepackage{graphicx}
\usepackage{multirow}
\usepackage{booktabs}
\usepackage{pgf-pie}
\usepackage[T1]{fontenc}
\usepackage{markdown}
\usepackage{amssymb}
\usepackage{pifont}

\usepackage{amsmath}

\newcommand{\cmark}{\ding{51}}%
\newcommand{\xmark}{\ding{55}}%

\usepackage[utf8]{inputenc}

\usepackage{microtype}
\usepackage{xcolor}

\title{Using Textual Interface to Align External Knowledge for End-to-End\\  Task-Oriented Dialogue Systems}

\author{
    Qingyang Wu, 
    Deema Alnuhait,
    Derek Chen,
    Zhou Yu  \\
     Columbia University \\
    \texttt{\{qw2345,daa2182,dc3761,zy2461\}@columbia.edu} \\
}

\begin{document}

\maketitle

\begin{abstract}

Traditional end-to-end task-oriented dialogue systems have been built with a modularized design.
However, such design often causes misalignment between the agent response and external knowledge, due to inadequate representation of information.
Furthermore, its evaluation metrics emphasize assessing the agent's pre-lexicalization response, neglecting the quality of the completed response.
In this work, we propose a novel paradigm that uses a textual interface to align external knowledge and eliminate redundant processes.
We demonstrate our paradigm in practice through MultiWOZ-Remake, including an interactive textual interface built for the MultiWOZ database and a correspondingly re-processed dataset.
We train an end-to-end dialogue system to evaluate this new dataset.
The experimental results show that our approach generates more natural final responses and achieves a greater task success rate compared to the previous models.
\end{abstract}







\section{Introduction}


\begin{figure*}[t]
    \centering
    \includegraphics[width=1.0\textwidth]{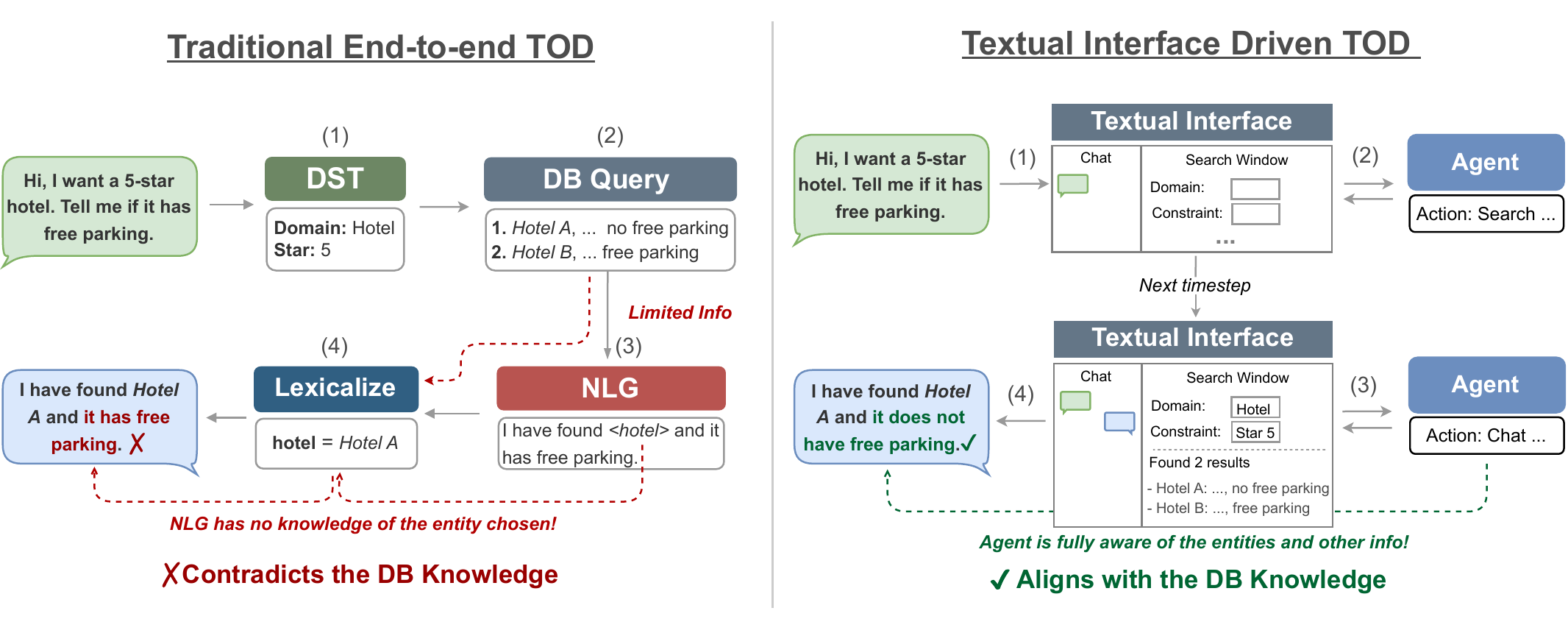}
    \caption{A comparative illustration of Traditional End-to-end TOD systems versus Textual Interface Driven TOD systems.
    This example highlights how the traditional pipeline may induce misalignment between the generated response and the corresponding database entity. Meanwhile, the textual interface in our pipeline demonstrates its effectiveness as a front-end for superior knowledge representation.
    More details are in Section~\ref{sec:methods}.}
    \label{fig:comparison}
\end{figure*}



Task-oriented dialogue systems have been extensively studied for various applications that involve natural language interactions between users and machines. 
These systems are designed to accomplish specific tasks, such as booking a hotel, ordering food, or scheduling an appointment.
The traditional paradigm for building such systems is to use a modularized design \cite{DBLP:conf/icassp/PieracciniTGGLL92,DBLP:conf/slt/Young06,DBLP:journals/pieee/YoungGTW13,DBLP:conf/sigdial/ZhaoE16}, where a dialogue state is maintained across modules to track the progress of the conversation and to interact with external databases. 
It generally incorporates Dialogue State Tracking (DST), database query or API calls, Natural Language Generation (NLG), and lexicalization to create the final system response.


However, the traditional modularized paradigm faces several limitations.
Firstly, it struggles to represent and integrate external knowledge effectively, as the modules operate independently, without a common knowledge grounding.
Secondly, the traditional paradigm heavily relies on lexicalization and delexicalization, resulting in annotations that are rigid and exhibit inconsistencies.
Moreover, current evaluation metrics primarily focus on assessing the agent's pre-lexicalization response, at the expense of neglecting the performance of the system as a whole, which compromises the end user experience.
As a result, this modularized design becomes a significant impediment in developing more effective end-to-end task-oriented dialogue systems.


To address these limitations, we propose a new task-oriented dialogue system paradigm that is Textual Interface Driven (TID) to better represent external knowledge and coordinate the interactions from the agent.
We instantiate our proposal using the existing MultiWOZ \cite{DBLP:conf/emnlp/BudzianowskiWTC18} dataset in order to highlight the differences.
As the original MultiWOZ dataset only contains limited annotations collected for the traditional paradigm,
we re-process it into MultiWOZ-remake by transforming the annotations into interface states and agent actions.
This new dataset simulates agent interactions over the textual interface, ensuring complete alignment of external knowledge representation with agent responses.
We also build an end-to-end dialogue agent for this dataset to demonstrate the effectiveness of our proposed paradigm.
%



In our experiments, we expose the problem of evaluating delexicalized responses with the commonly used metrics of `Inform' and `Success'. Instead, we evaluate the final lexicalized responses with BLEU to better reflect the performance of the end-to-end system.
To more thoroughly assess the system, we conduct a comprehensive human evaluation. 
Compared against strong baselines, our system generates more natural responses and achieves a higher task success rate, thereby showcasing the superiority of our proposal.

    
    


\section{Related Work}


The most common task-oriented dialogue paradigm is the dialogue state paradigm, or slot-filling paradigm \cite{DBLP:conf/icassp/PieracciniTGGLL92,DBLP:conf/slt/Young06,DBLP:journals/pieee/YoungGTW13,DBLP:conf/sigdial/ZhaoE16}.
It typically consists of several modular components, including a natural language understanding module that extracts user intents and relevant information for the task \cite{hashemi2016query, shi-etal-2016-deep},  dialogue state tracking module which tracks the current state of the conversation \cite{kim2017fourth}, a dialogue policy module for learning dialogue acts, and a natural language generation module to generate the system response.

The MultiWOZ dataset \cite{budzianowski2018multiwoz} extends this paradigm by providing comprehensive annotations for building different dialogue systems \cite{DBLP:conf/acl/WuMHXSF19,DBLP:conf/eacl/WuZLY21,gu-etal-2021-pral,DBLP:conf/emnlp/Lee21,DBLP:journals/corr/abs-2005-00796,https://doi.org/10.48550/arxiv.2211.16773}.
However, the traditional task-oriented dialogue system paradigm has limitations in effectively representing the external knowledge.
In this work, we address these limitations and remake MultiWOZ with our proposed paradigm.

\section{Textual Interface-Driven TOD}
\label{sec:methods}


Our textual interface-driven (TID) approach effectively circumvents the limitations of the traditional modularized design, where each module requires a specific schema for inter-module communication, leading to ineffective knowledge representation and error propagation throughout the conversation. 
Conversely, our model leverages a \textit{unified} textual interface, serving as a precise and comprehensive front-end for knowledge representation.

In the following subsections, we initially outline the implementation of the textual interface using the document tree. 
We then present a comparative illustration between the traditional paradigm and our proposed one. 
Finally, we show the construction of an end-to-end dialogue agent for our interface.


\subsection{Interface with Document Tree}
To better represent information, we utilize a virtual document tree to implement the textual interface, similar to the document object model \cite{keith2006dom} employed in HTML, where each node can represent part of the document such as a title, text span, or a list.
This approach captures the document's structure as a hierarchical, tree-like object.
It also helps to separate the presentation of content from its underlying structure and behavior, making it easier to update the interface representation.
To preserve formatting and structural information, we further render the document tree into Markdown. 
Markdown is a lightweight markup language that is used for formatting text. 
It provides a simple and easy-to-use syntax for creating headings, lists, and other elements, and it is designed to be easy to read and write \cite{mailund2019introducing}.
This rendered Markdown text will serve as the state representation as the dialogue system's inputs.
For more details and illustrations, please see Appendix~\ref{sec:multiwoz_dom}.

\subsection{Comparative Illustration}

Figure~\ref{fig:comparison} provides a comparative illustration of the traditional paradigm versus our proposed paradigm.
In most of the traditional TOD implementations, there are four main stages. Initially, a user's input is processed by (1) a dialogue state tracking (DST) module, extracting the user's intentions and beliefs. 
Subsequently, (2) a database (DB) query is conducted using the extracted intents and belief states. 
Next, (3) a natural language generation (NLG) module creates a delexicalized response, exemplified in the figure. 
Finally, (4) in the next step, the placeholders in the delexicalized response are replaced with actual entities information derived from the database query.

However, such a design spreads dialogue states across the modules, causing difficulties in syncing database information with the actual generated response. 
In the provided example, the system's response inaccurately reflects that 'Hotel A' has free parking, yet the NLG module is unaware of the specific entity chosen.
Similar misalignment can occur with the DST module, especially when managing booking requests, as it may lack knowledge of the previously selected entity.

In contrast, our interface-driven paradigm avoids misalignment by having a \textit{shared} textual interface to coordinate all the information flow between the user and agent.
(1) A user's utterance updates the interface's state. 
The agent then determines the next action. 
For instance, the agent may execute a search on the interface (2), which updates its state. 
At the next timestep, the agent choose to 'Chat' through the interface (3), 
and the final response is delivered to the user (4).
The agent is fully aware of the entity displayed on the interface and can generate a consistent and cogent response based on its selected entity.

\subsection{End-to-End TOD Agent}

\begin{figure}[t]
    \centering
    \includegraphics[width=0.45\textwidth]{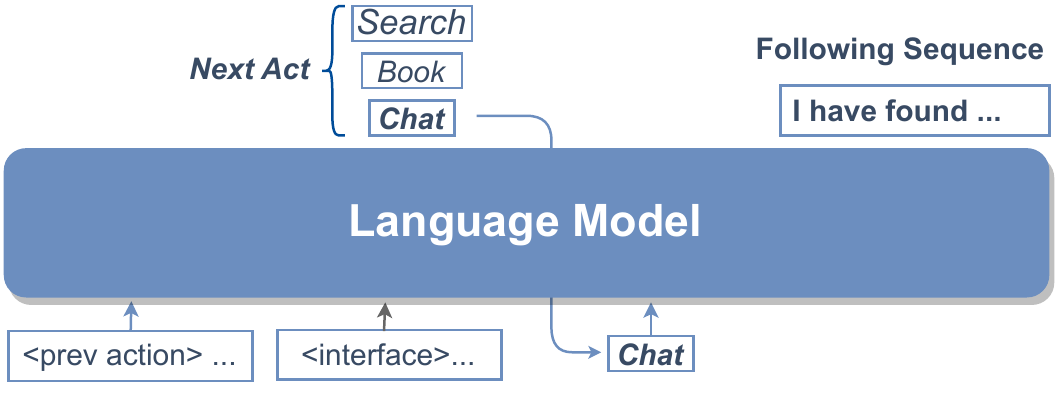}
    \caption{In our end-to-end model, the agent initially predicts the next action, and then generates the following sequence based on that action.}
    \label{fig:my_label}
\end{figure}

To interact with the textual interface, we build a model that is compatible with most task-oriented dialogue datasets.
The input context contains the previous action and the current textual interface state.
The model needs to first predict the next action.
It includes three main next actions: \textit{``Chat''},  \textit{``Book''} and \textit{``Search''}.
Then, the predicted action is fed back to the model.
\textit{``Chat''} continues to output the generated sequence to the chat window, while \textit{``Search''} and \textit{``Book''} updates the search constraint or booking information displayed in the search window with the following generated sequence.
This setting is compatible with different language models including encoder-decoder models. 





\section{MultiWOZ Remake}

We remake the existing MultiWOZ dataset \cite{budzianowski2018multiwoz} to showcase the usefulness of our proposed paradigm. 
We implemented a textual interface to interact with the database and re-processed the dataset accordingly. 

\subsection{MultiWOZ Interface}

We developed a textual interface that interacts with the MultiWOZ's database, which follows the interface-driven paradigm and utilizes the document tree design.
For each of the seven domains present in MultiWOZ, we design a different sub-section in the interface based on the query domain.
This interface defines the front-end and the back-end functions.
The front-end displays necessary details such as query domain, constraints, the number of entities found, and booking status.
It also displays a truncated list of presently searched entities for agent selection.
Meanwhile, the back-end handles SQL search calls, utilizing current and prior constraints entered into the interface to identify appropriate entities, and it also verifies booking availability and, if successful, returns a reference number.
An example can be found in Appendix~\ref{sec:multiwoz_dom}.



\subsection{Data Re-Processing}

The original MultiWOZ dataset did not record the selected entities during the conversation, leading to a misalignment between the interface representation and the actual response. 
Therefore, we need to re-process the dataset to replay the agent's actions on the interface, thereby ensuring alignment between the selected entity and the interface representation.

We use MultiWOZ 2.2 \cite{zang-etal-2020-multiwoz} as it provides necessary annotations to help us re-process the dataset.
Specifically, we track entities from previous dialogue history, ensuring alignment between the query domain, search constraints, and selected entities. 
In particular, the entity chosen during booking should correspond with the actual booked entity. 
Detailed reprocessing steps can be found in Appendix~\ref{sec:reprocess_details}.

Note that due to annotation errors and human inconsistency, 23\% of dialogues have issues tracking entities throughout the conversation, mainly occurring in multi-domain dialogues.
To minimize noise, these dialogues are excluded from training. For more details regarding these inconsistencies, please refer to Appendix~\ref{sec:reprocessing_inconsistency}.

\section{Experiments}
We conducted several experiments on the MultiWOZ test set to evaluate our end-to-end dialogue agent.
We tested different back-bone encoder-decoder models including BART \cite{DBLP:journals/corr/abs-1910-13461}, T5 \cite{DBLP:journals/jmlr/RaffelSRLNMZLL20}, and GODEL \cite{https://doi.org/10.48550/arxiv.2206.11309} to compare with previous models, by fine-tuning them on the re-processed MultiWOZ Remake training set to compare with the baselines. 

In this section, we first show the problems for the evaluation metrics: Inform and Success, which are widely used in the task-oriented response generation task. 
Then, we show the automatic evaluation results with a more direct metric.
In the end, we show human evaluation results to better demonstrate the performance of our approach.

\subsection{Problems of Inform and Success}
\label{sec:evaluation_metrics}


\begin{table}[t]
    \centering
    \begin{subtable}[h]{0.243\textwidth}
        \centering
        \resizebox{1.0\textwidth}{!}{
            \begin{tabular}{l | c c c c}
            \toprule
            \textbf{Model} & \textbf{Inform} & \textbf{Success} \\
            \midrule
            MTTOD & 85.9 & 76.5 \\
            GALAXY & 85.4 & 75.7 \\
            Mars & 88.9 & 78.0 \\
            KRLS & 89.2 & 80.3 \\
            \midrule
            fixed response* & \textbf{89.2} & \textbf{88.9} \\
            \bottomrule
           \end{tabular}
       }
       \caption{End-to-end models}
       \label{tab:1}
    \end{subtable}
    \hfill
    \begin{subtable}[h]{0.23\textwidth}
        \centering
        \resizebox{1.0\textwidth}{!} {
            \begin{tabular}{l | c c c c}
            \toprule
            \textbf{Model} & \textbf{Inform} & \textbf{Success} \\
            \midrule
            Gold (human) & 93.7 & 90.9 \\
            \midrule
            KRLS & 93.1 & 83.7 \\
            MarCO & \textbf{94.5} & 87.2 \\
            GALAXY & 92.8 & 83.5 \\
            \midrule
            fixed response* & 92.5 & \textbf{92.3} \\
            \bottomrule
           \end{tabular}
       }
        \caption{Policy optimization}
        \label{tab:2}
     \end{subtable}
    \caption{Inform and Success scores. * a fixed response is used: ``[value\_name] [value\_phone] [value\_address] [value\_postcode] [value\_reference] [value\_id]''}
    \label{tab:metrics_limitation}
\end{table}

Task-oriented dialogue systems often use Inform and Success to evaluate the quality of response generation.
However, they are designed for delexicalized responses like ``[value\_name] is a restaurant...'', and it needs further lexicalization process to fill in the placeholders like ``[value\_name]''.
As a result, they are not reflecting the real quality of the final response.

Furthermore, we question the validity of the current Inform and Success metrics implementation.
They check the cumulative belief states for placeholders and whether the response contains a reference to calculate the scores.
Consequently, a model that generates more placeholders achieves a deceptively better performance.

To illustrate this, we use the same fixed response ``[value\_name] [value\_phone] [value\_address] [value\_postcode] [value\_reference] [value\_id]'' for every turn when evaluating on the standardized evaluation script \cite{DBLP:journals/corr/abs-2106-05555} to report the performance. 
We compare both end-to-end and policy optimization models.
In the end-to-end setting, we use the dialogue state prediction from Mars \cite{DBLP:journals/corr/abs-2210-08917}.

Table~\ref{tab:metrics_limitation} shows the results with this fixed response. 
Surprisingly, the fixed response achieves state-of-art performance on the Inform and Success score compared to the baseline models.
It is questionable whether Inform and Success can measure the true performance of the system, and they may mislead the existing models.
We urge future researchers to stop reporting Inform and Success until a better evaluation metric is proposed.

\begin{table}[t]
    \centering
    \resizebox{0.38\textwidth}{!}{
    \begin{tabular}{l |  c c c }
        \toprule
        \textbf{Model} & \textbf{Backbone} & \textbf{\#Parameters} & \textbf{BLEU} \\
        \midrule
        HDSA & $\textrm{BERT}_{base}$ &  110M & 11.87 \\
        \midrule
        MTTOD & $\textrm{GODEL}_{base}$ & 360M & 13.83 \\
         & $\textrm{GODEL}_{large}$  & 1.2B  & 13.06 \\
        \midrule
        GALAXY & $\textrm{UniLM}_{base}$ & 55M & 13.71 \\
        \midrule
        Mars & $\textrm{T5}_{base}$ & 220M  & 13.58 \\
        \midrule
        Remake & $\textrm{BART}_{base}$ & 140M & 15.87 \\
        & $\textrm{BART}_{large}$ & 406M & 15.82 \\
        & $\textrm{T5}_{base}$ & 220M & 15.27 \\
        & $\textrm{T5}_{large}$ & 770M & 16.66 \\
        & $\textrm{GODEL}_{base}$ & 220M & 16.55 \\
        & $\textrm{GODEL}_{large}$ & 770M & \textbf{16.92} \\
        \bottomrule
    \end{tabular}
    }
    \caption{BLEU results for lexicalized responses.}
    \label{tab:automatic_evaluation}
\end{table}

\subsection{Automatic Evaluation}

We use the sentence level sacreBLEU \cite{post-2018-call} to evaluate the performance of various task-oriented dialogue systems.
We compare Remake with strong baselines including HDSA \cite{DBLP:conf/acl/ChenCQYW19}, MTTOD \cite{DBLP:conf/emnlp/Lee21}, GALAXY \cite{DBLP:conf/aaai/HeDZWCLJYHSSL22}, and Mars \cite{DBLP:journals/corr/abs-2210-08917}.
Note that we evaluate the quality of final lexicalized responses.
We use the lexicalization script provided by \citet{DBLP:journals/corr/abs-2005-00796} to fill in the placeholders for the baselines' outputs.
We use the policy optimization setting for all models.

As shown in Table~\ref{tab:automatic_evaluation}, Remake models with the new paradigm achieve better performance than the baseline models (HDSA, MTTOD, GALAXY, and Mars)with the traditional dialog state paradigm. 
Especially, although HDSA \cite{DBLP:conf/acl/ChenCQYW19} has the best reported BLEU score performance with delexicalized responses, it gets worse performance after lexicalization.
This observation suggests that our paradigm model can greatly improve the quality of final lexicalized responses. 

\subsection{Human Evaluation}

As mentioned in Section~\ref{sec:evaluation_metrics}, automatic evaluation metrics can be misleading.
Thus, we conduct human evaluations to better evaluate the performance improvement of our model Remake compared to MTTOD, as it is the best-performing baseline.

We hire a human worker to talk with each model for 21 whole conversations using the same goal instructions from the MultiWOZ dataset.
On average, each conversation finishes in ten turns.
Then, the worker rates the models in terms of two metrics: ``Goal Success'' and ``Coherence''. 
``Goal Success'' measures if the system can successfully satisfy the user's goal without given any information contradicting to the database.
``Coherence'' measures if the system responses are coherent and human-like.

\begin{table}[t]
    \centering
    \resizebox{0.38\textwidth}{!}{
    \begin{tabular}{l c c}
        \toprule
        \multicolumn{1}{c}{\textbf{Model}} & \multicolumn{1}{c}{\textbf{Backbone}}  & \textbf{Goal Success (\%)} \\
        \midrule
        MTTOD & $\textrm{GODEL}_{base}$ & 47.6 \\
        & $\textrm{GODEL}_{large}$ & 38.1 \\
        \midrule
        Remake & $\textrm{GODEL}_{large}$ & \textbf{90.5} \\
        \bottomrule
    \end{tabular}
    }
    \caption{Human Evaluation for Goal Success.}
    \label{tab:human_evaluation_goal}
\end{table}

\begin{table}[t]
    \centering
    \resizebox{0.39\textwidth}{!}{
    \begin{tabular}{l  c c c}
        \toprule
         \multicolumn{1}{c}{\textbf{Comparison}}  & \textbf{Win } & \textbf{Lose} & \textbf{Tie} \\
        \midrule
        $\textrm{Remake}$ vs. $\textrm{MTTOD}_{base}$ & 57.1\%  & 0.0\% & 42.9\%   \\
        $\textrm{Remake}$ vs. $\textrm{MTTOD}_{large}$ & 52.4\%  & 4.8\%  & 42.8\%  \\
        \bottomrule
    \end{tabular}
    }
    \caption{Human evaluation for coherence.}
    \label{tab:human_evaluation_coherence}
\end{table}

Table~\ref{tab:human_evaluation_goal} shows the human evaluation results for goal success.
The Remake model demonstrates a significantly higher level of accuracy (p < 0.01), achieving 90.5\% goal success, compared to the 47.6\% accuracy of the MTTOD model.
This improvement suggests that the use of an interface can help the system reduce hallucinations and better satisfy the user's request.

Table~\ref{tab:human_evaluation_coherence} shows the human evaluation results for coherence. 
``Win'' indicates that the dialogue looks more coherent, whereas ``Lose'' means the opposite. 
Remake is significantly more coherent than MTTOD (p < 0.01 with paired t-test).
We observe that, in the context of an entire conversation, MTTOD struggles to maintain entity consistency, resulting in incoherent dialogues.

\begin{table}[t]
    \centering
    \resizebox{0.4\textwidth}{!}{
    \begin{tabular}{l | c | c c c }
        \toprule
        \textbf{Backbone} & \textbf{Prev Act} & \textbf{Next Act} & \textbf{Search
        } & \textbf{BLEU} \\
        \midrule
        $\textrm{BART}_{base}$  & \xmark  & 62.0 & 3.8 & 11.24 \\
        $\textrm{BART}_{large}$ & \xmark & 57.8 & 4.2 & 13.51 \\
        \midrule
        $\textrm{BART}_{base}$ & \cmark & 92.1 & 77.4  & 15.87 \\
        $\textrm{BART}_{large}$ & \cmark & \textbf{95.2} & 77.5 & 15.82 \\
        \midrule
        $\textrm{GODEL}_{base}$ & \cmark & 95.0 & 76.6 & 16.55 \\
        $\textrm{GODEL}_{large}$ & \cmark & 95.1 & \textbf{79.7} & \textbf{16.92} \\
        \bottomrule
    \end{tabular}
    }
    \caption{Action prediction results and ablation studies. 
    ``Next Act'' means the next action's prediction accuracy. ``Search`` means the search query accuracy.}
    \label{tab:ablation_studies}
\end{table}


\subsection{Ablation studies}

We perform ablations of our model for the Next Act prediction accuracy and the Search prediction accuracy.
The search query accuracy measures if the system generates the correct sequence when searching the database when performing \textit{``Search``} action.
Table~\ref{tab:ablation_studies} shows the final results.
The backbone with GODEL-large achieves the best overall performance.
Note that without the previous action in the context, the model is unaware of its previous action and performs not well, which suggests the importance of the history state in our paradigm.

We also conducted the error analysis for the  wrongly predicted search action with three categories prediction error, annotation error, and ignore type. 
More details can be found in Appendix~\ref{sec:search_error}.

\section{Conclusion}
In conclusion, we have proposed a novel textual interface driven paradigm for building task-oriented dialogue systems. 
The traditional paradigm struggles to effectively represent external knowledge across different modules. 
By replacing the dialogue state with an interactive textual interface, our system allows for more efficient and accurate coordination of the data flowing between the user, the agent, and the knowledge base.
We have instantiated this in practice by presenting MultiWOZ-Remake, an interactive interface for the MultiWOZ database, and a corresponding dataset. 
Experimental results show that our system in this new paradigm generates more natural responses and achieves a greater task success rate compared against the previous models.

\newpage





\bibliography{anthology,custom}
\bibliographystyle{acl_natbib}

\begin{figure*}[t]
    \centering
    \includegraphics[width=0.92\textwidth]{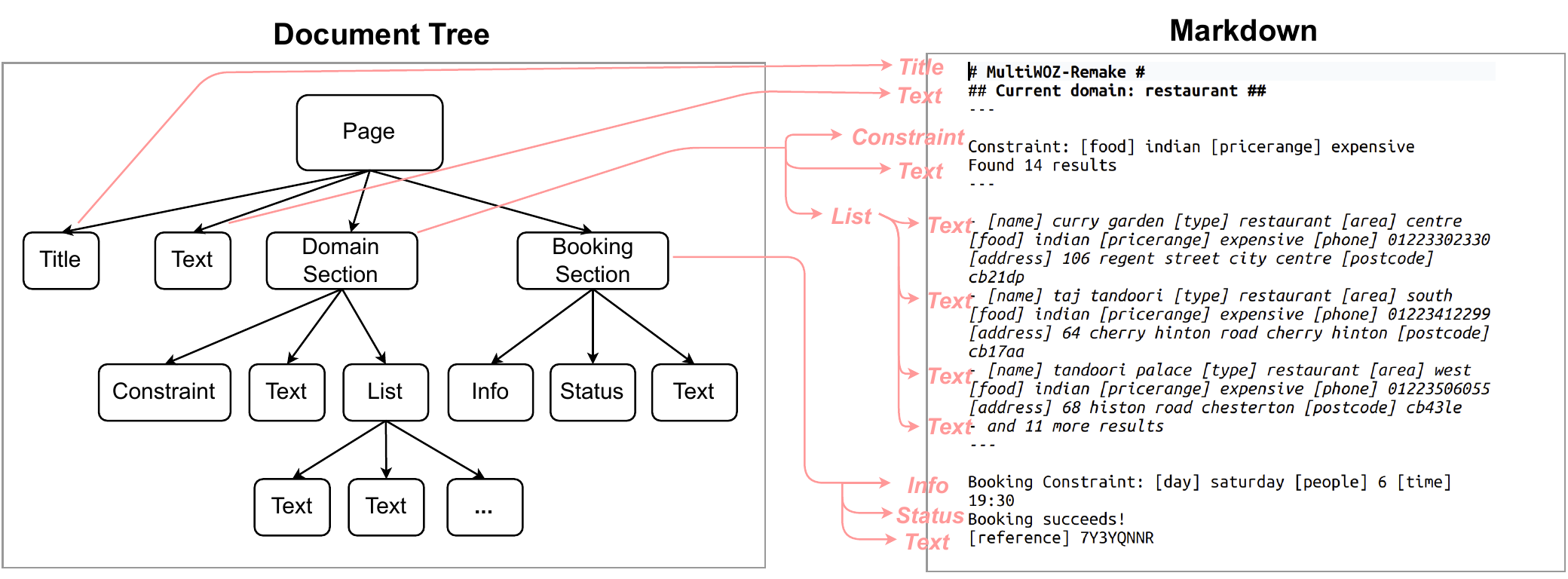}
    \caption{An example of rendering the Document Tree representation into the textual representation in Markdown. Document tree simplifies the manipulation of the dynamic elements in the interface, while Markdown is to display richly formatted text. This approach provides both the flexibility and comprehensibility for the interface.}  
    \label{fig:interface_markdown}
\end{figure*}

\newpage
\appendix

\section{MultiWOZ Remake}

\subsection{Document Tree and Textual Interface}
\label{sec:multiwoz_dom}

Figure~\ref{fig:interface_markdown} provides how MultiWOZ's interface is implemented and it highlights the transformation between the Document Tree and Markdown. 
This illustration provides a visual representation of how the different components of the interface correspond to the Document Tree and Markdown.
The right part of figure~\ref{fig:interface_markdown} shows an example of the interface in the restaurant domain.
It highlights the flexibility of using Document Tree to manipulate dynamic elements such as ``Domain Section'', ``Booking Section'', or ``List'', and then it can be rendered into the Markdown text that is comprehensible for both humans and language models.

The interface can be interacted with by providing a command \textit{``[domain] [slot] value''} or \textit{``[booking] [slot] value''}.
\textit{``[slot] value''} is optional if only domain switch is performed.
The command would update the internal constraint for querying the database and refresh the dynamic elements showing in the interface.
For example, in Figure~\ref{fig:interface_markdown}, the interface state can be reached by performing two actions: \textit{``[restaurant] [food] indian [pricerange] expensive''} and \textit{``[booking] [day] saturday [people] 6 [time] 19:30''}. 
It is important to note that there can be multiple different paths of actions to reach the same interface state, allowing for flexibility in how the agent interacts with the interface.

\subsection{Re-processing details}
\label{sec:reprocess_details}

The proposed interface for the MultiWOZ database implements two types of search commands: searching with constraints, and booking using provided information.
To simplify the complexity of the command, the interface uses the incremental belief state between two turns as the query action.
In the back-end, a cumulative belief state is used to perform the actual SQL search.
Additionally, when multiple domains are involved in a single turn, we divide this turn into multiple actions to ensure completeness.

For the booking functionality, we used multiple sources of information, including the span annotation for the booking reference number, dialogue as annotations to provide booking status, and information from belief states to determine whether a booking takes place at the current turn. 
Additionally, we aligned the interface's representation with the recorded booking outcome, whether it is a success or failure. 
Therefore, the interface can correctly display the booking status when handling the booking action.

The details of the listed database results are truncated to avoid a long context.
However, it is possible that the mentioned entities in the training data's response are truncated from the list.
To avoid this situation, we re-arranged the database search results so that the mentioned entities are always shown in the interface display, which minimizes the hallucination with the correct entity grounding.

\subsection{Re-Processing Inconsistency}
\label{sec:reprocessing_inconsistency}


It is important to note that there can be inconsistency between the training data and the re-processed data by replaying the trajectories on the interface.
If some entities in the response cannot be inferred from the history context, we recognize it as a inconsistent dialogue.
This normally happens due to the annotation errors \cite{https://doi.org/10.48550/arxiv.2212.02745} in MultiWOZ, or the complex scenarios when multiple domains are involved.

There are 2373 out of 10438 dialogues that are potentially inconsistent during data processing.
Then, we randomly sampled 250 dialogues and manually classify them to check the consistency. 
We observed that single-domain dialogues have 74\% consistency and multiple-domain dialogues have 43\% consistency, suggesting that multiple-domain dialogues are more complex.
Also, we found that the consistency is relatively disproportional to the length of turns. 

\definecolor{orange}{HTML}{FF7F00}

\begin{figure*}[h]
        \centering
    \includegraphics[width=0.98\textwidth]{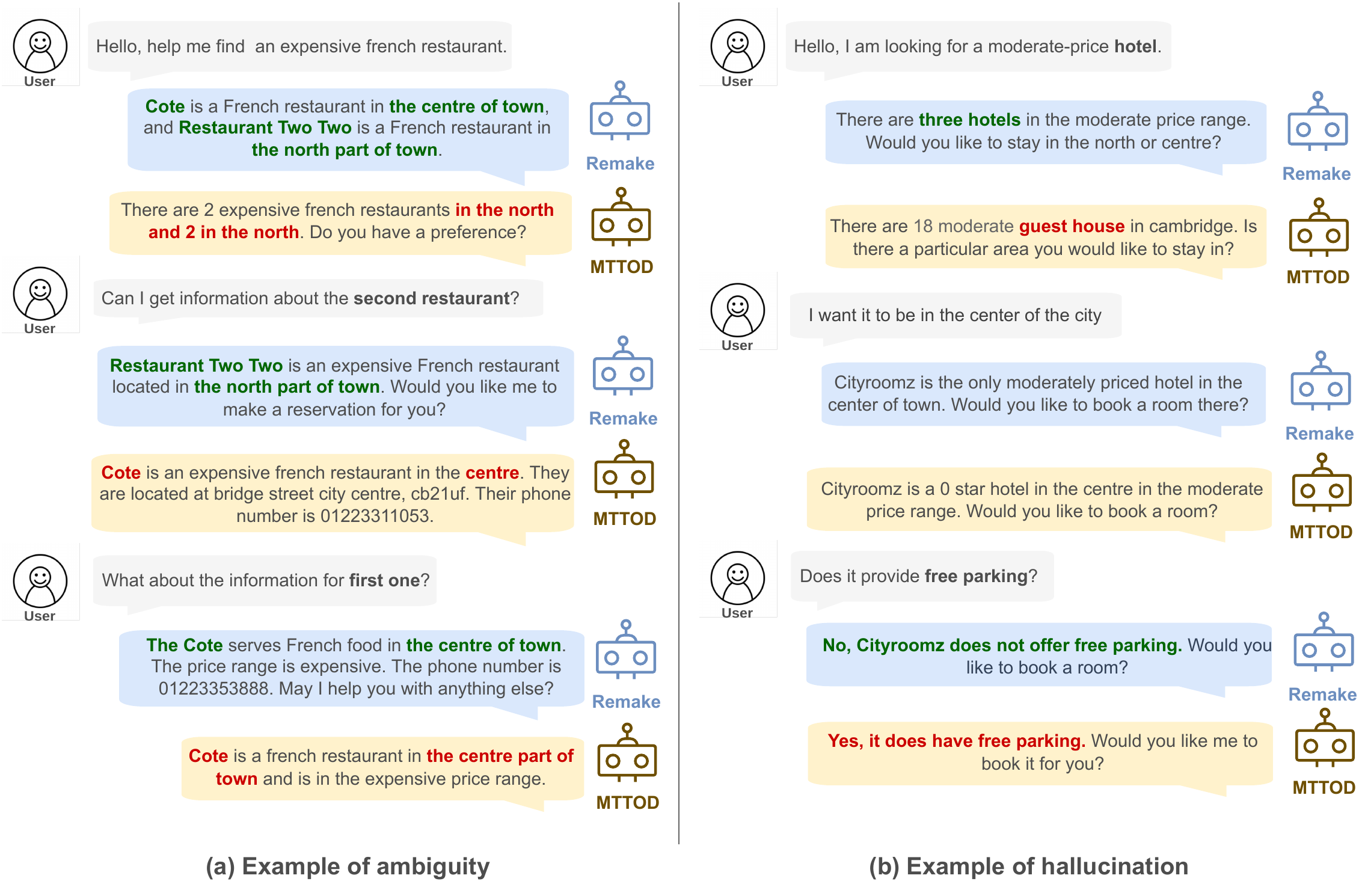}
    \caption{Dialogue examples that address the problems of ambiguity and hallucination. We compare Remake and MTTOD's outputs.
    \textcolor[HTML]{006600}{\textbf{Green}} represents the correctly generated spans. \textcolor[HTML]{CC0000}{\textbf{Red}} represents the incorrect ones.
    }
    \label{fig:dialog_example}
\end{figure*}

\section{Search Error Analysis}
\label{sec:search_error}
For error analysis, we divided the search prediction errors into three categories: Type I: Prediction Error where the model makes a wrong prediction, Type II: Annotation Error; and Ignore: Error which can be ignored. 


\begin{table}[h]
    \centering
    \resizebox{0.45\textwidth}{!}{
    \begin{tabular}{l | c }
        \toprule
        \textbf{Error Types} & \textbf{Percentage} \\
        \midrule
        Type 1: Prediction Error & 40.0\% \\
        Type 2: Annotation Error (Labeling) & 6.0\% \\
        Type 2: Annotation Error (Discourse) & 2.0\% \\
        Ignore & 52.0\%  \\
        \bottomrule
    \end{tabular}
    }
    \caption{Percentage of different errors.}
    \label{tab:dst_errors}
\end{table}

For the prediction error, the common mistake is forgetting to predict one of the intents requested by the user.
Sometimes, this can be due to mispredicting attributes that require reasoning.
Also, searching for ``train'' domain requires attributes like destination and departure to be all revealed.
For example, if the user says ``I want to book the restaurant for tomorrow.'', then the agent needs to transfer that into the actual value represented in the database.
For the annotation errors, we further divided them into labeling errors, ontology and inconsistencies, and discourse errors as suggested by \citet{https://doi.org/10.48550/arxiv.2212.02745}. The labeling errors occur when the states are under-labeled or over-labeled while discourse attributes are when the dialogues show occurrences of inconsistency or incoherence.

We randomly select 50 errors and classify them into those categories.
Table \ref{tab:dst_errors} shows the results.
We can observe that most errors can be ignored. 
However, the model still accounts for a large portion of the errors, suggesting that the model needs further improvement in terms of search.



\section{Case Studies}

Figure~\ref{fig:dialog_example} shows two dialogue examples chatting with Remake and MTTOD, respectively.
It demonstrates the common problems of the traditional dialog state paradigm.
The first problem is handling ambiguity in the user's utterance, which is previously studied by \citet{DBLP:journals/corr/abs-2112-08351}.
MTTOD cannot handle such requests very well as the lexicalization process involves no understanding.
The same situation can happen when the user says ``what about another restaurant?''

Another type of problem is hallucination. 
Models like MTTOD often use the number of the returned database results to represent the grounding of the database.
As a result, it cannot handle complex questions from the user.
In this example, ``Cityroomz'' does not offer free parking at all, but MTTOD hallucinates to provide the wrong information to the user.
It suggests the necessity of using our paradigm to provide knowledge grounding for the model to avoid this case.










\end{document}